\documentclass[journal]{IEEEtran}
\usepackage{multirow}
\usepackage{bm}
\usepackage{color}
\usepackage{epsfig}
\usepackage{graphicx}
\usepackage{amsmath}
\usepackage{amssymb}
\usepackage{algorithm} 
\usepackage{algorithmic} 
\usepackage{amsthm}
\usepackage{array}
\usepackage{colortbl}
\usepackage{bbding}
\usepackage{url}
\usepackage{hyperref}

\theoremstyle{remark}

\newcommand{\eg}{\textit{e.g.~}}
\newcommand{\ie}{\textit{i.e.~}}

\newcommand{\etc}{\textit{etc.~}}

\newcommand{\etal}{\textit{et al.~}}

\ifCLASSINFOpdf
\else
\fi

\begin{document}
	%
	\title{Geometric Property Guided Semantic Analysis \\ of 3D Point Clouds}
	\title{Auxiliary Geometric Learning on Point Clouds}
	\title{Improving Point Cloud Analysis by Auxiliary Deep Regression on Geometric Properties}
    \title{Improving Semantic Analysis on Point Sets by Auxiliary Regression on Geometric Properties}	
    \title{Deep Geometric Learning with Auxiliary Regression on Surface Properties}
    \title{Improving Semantic Analysis on Point Clouds via Auxiliary Supervision of Geometric Properties}
    \title{Improving Semantic Analysis on Point Clouds via Auxiliary Supervision of Local Geometric Priors}

\author{Lulu Tang$^{*}$, Ke Chen$^{*}$,~\IEEEmembership{Member,~IEEE,}~Chaozheng Wu, Yu Hong, Kui Jia$^\dagger$,~\IEEEmembership{Member,~IEEE,}~and Zhi-Xin Yang$^\dagger$,~\IEEEmembership{Member,~IEEE}
\thanks{$^*$equal contribution to this work; $^\dagger$corresponding author.}

\thanks{ This work was funded in part by the Science and Technology Development Fund, Macau SAR (File no. SKL-IOTSC-2018-2020, 0018/2019/AKP, 0008/2019/AGJ, and FDCT/194/2017/A3), in part by the National Natural Science Foundation of China (Grant No.: 61771201, 61902131),  in part by the Program for Guangdong Introducing Innovative and Enterpreneurial Teams (Grant No.: 2017ZT07X183), in part by the Fundamental Research Funds for the Central Universities (Grant No.: 2019MS022), and in part by the University of Macau (Grant No.: MYRG2018-00248-FST and MYRG2019-0137-FST).}
\thanks{L. Tang and Z. Yang  are with the State Key Laboratory of Internet of Things for Smart City and Department of Electromechanical Engineering, University of Macau, Macau SAR, China, E-mails: lulu.tang@connect.um.edu.mo; zxyang@um.edu.mo.}
\thanks{K. Chen, C. Wu, Y. Hong and K. Jia are with the School of Electronic and Information Engineering, South China University of Technology, Guangzhou 510641, China. K. Chen is also with the Peng Cheng Laboratory, Shenzhen 518005, China.}
}
	
\markboth{Journal of \LaTeX\ Class Files,~Vol.~X, No.~X, January~2020}%
{Shell \MakeLowercase{\textit{et al.}}: Bare Demo of IEEEtran.cls for IEEE Journals}	

\maketitle

\begin{abstract}
Existing deep learning algorithms for point cloud analysis mainly concern discovering semantic patterns from global configuration of local geometries in a supervised learning manner. 
However, very few explore geometric properties revealing local surface manifolds embedded in 3D Euclidean space to discriminate semantic classes or object parts as additional supervision signals.
This paper is the first attempt to propose a unique multi-task geometric learning network to improve semantic analysis by auxiliary geometric learning with local shape properties, which can be
either generated via physical computation from point clouds themselves as self-supervision signals or provided as privileged information.
Owing to explicitly encoding local shape manifolds in favor of semantic analysis, the proposed geometric self-supervised and privileged learning algorithms can achieve superior performance to their backbone baselines and other state-of-the-art methods, which are verified in the experiments on the popular benchmarks.	
\end{abstract}
	
\begin{IEEEkeywords}	
Geometric properties, point clouds, semantic analysis, self-supervised learning, privileged learning.
\end{IEEEkeywords}

\IEEEpeerreviewmaketitle
	
\section{\textbf{Introduction}}
	
	
Point clouds collecting a set of order-less points to represent 3D geometry of objects have been verified as a powerful shape representation in a number of recent works \cite{liu2015robotic,he2019geonet,qi2017pointnet,qi2017pointnet++,li2018pointcnn,wang2018dynamic,wen2019geometry,wang2019deep}. 
Semantic analysis on a point set aims to categorizing the points globally into semantic classes (\eg plane, chairs, mugs) \cite{qi2017pointnet,qi2017pointnet++,wang2018dynamic,wu2016learning,li2018so,yang2018foldingnet} or locally into object parts \cite{qi2017pointnet,wang2018dynamic,li2018so} according to their topological configuration. 
Such a problem plays a vital role in many applications, especially those demanding visual perception and interaction between machines and surrounding environment such as augmented reality, robotics and automatic driving. 
Semantic patterns of point clouds can be discovered from global configuration of local geometric patterns, but it is challenging to discover and exploit such local geometries due to inherently missing point-wise connectivity in their neighborhood.
	
	
A number of recent works have been proposed to feature learning on point sets, via either designing locally-connected convolutional/pooling layers on irregular non-Euclidean points such as PointNet \cite{qi2017pointnet}, PointCNN \cite{li2018pointcnn}, Dynamic Graph CNN (DGCNN) \cite{wang2018dynamic}, and GeoNet \cite{he2019geonet}, or hierarchically aggregating features revealing geometric patterns across scales, \eg PointNet++ \cite{qi2017pointnet++}, SO-Net \cite{li2018so}.
These existing methods in a supervised learning manner utilize pre-defined annotations to implicitly learn a global topology and local geometries sensitive to semantic classes. 
Very few work pays an attention to explicitly constraining 3D neural classifiers with auxiliary regressing onto local geometric properties.

\begin{figure}[t]
\centering \includegraphics[width=0.995\linewidth]{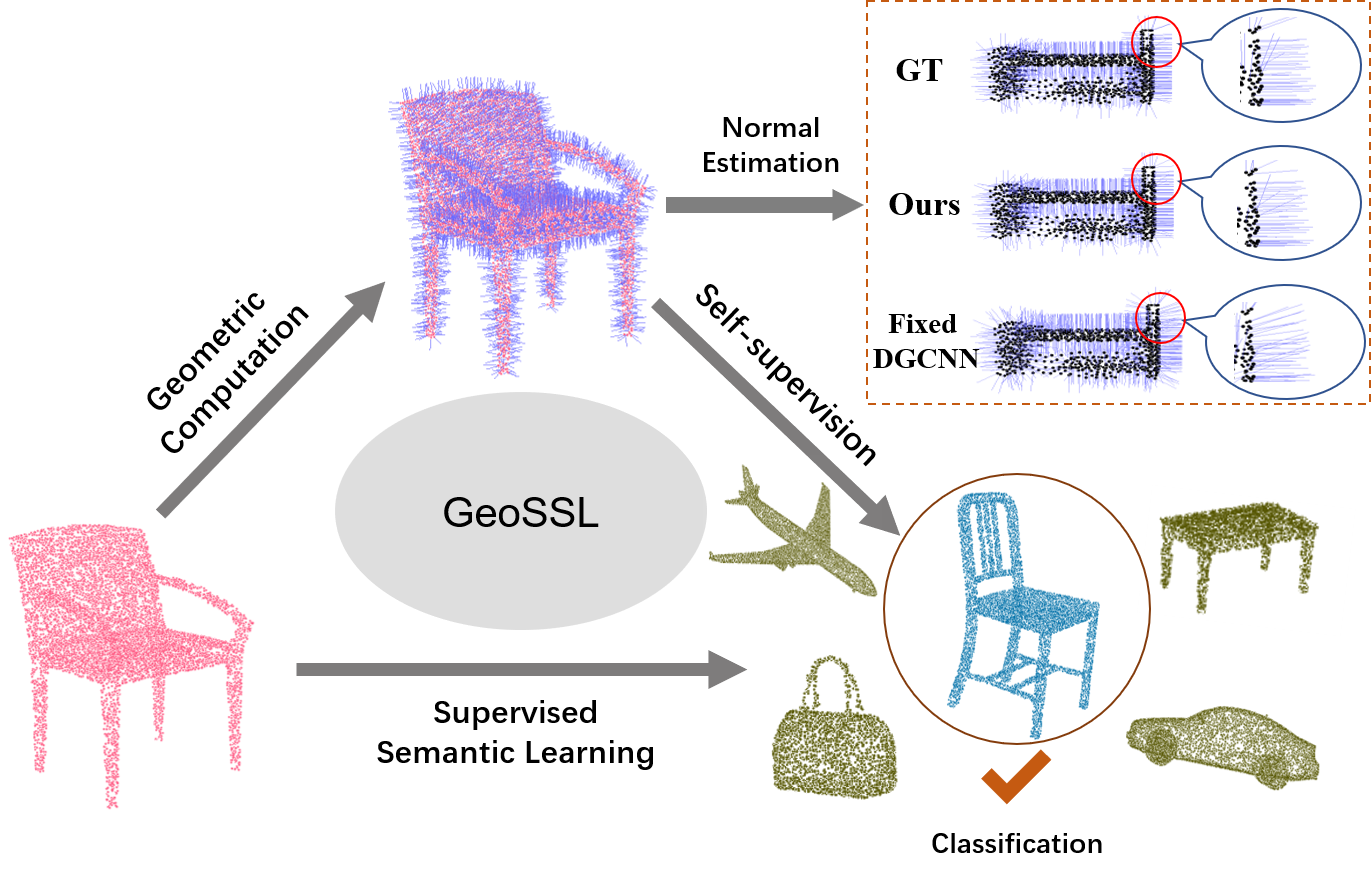}
\caption{A flow chart of the proposed geometric self-supervised learning (GeoSSL): Geometric properties generated by physical computation are considered as self-supervised signals to support supervised semantic shape analysis. Owing to additionally fitting geometric properties, the backbone methods (\eg PointNet++ \cite{qi2017pointnet++}, DGCNN \cite{wang2018dynamic} in our experiments) can be improved for semantic analysis on point sets.
}
\label{fig:intro}
\end{figure}
	

Local geometric properties such as point-wise normal vectors, curvatures, and tangent spaces \etc  are the primitive properties of local point groupings that reveal local geometric manifolds.
For example, for computing a normal of a point, the typical solution is to first fit a plane via a set of its nearest neighboring points and obtain the normal of the plane, which indicates point-wise geometric properties describing local connectivity across nearby points.
Some works \cite{guerrero2018pcpnet,ben2018nesti} design a deep network to directly estimate these geometric properties from point clouds.
However, local geometric properties can be freely obtained by physical computation with no price for additional efforts on manually annotation, especially for massively amounts of auxiliary data that are usually produced by computer aided design (CAD).
	
Point-wise geometric properties, in most of existing works \cite{qi2017pointnet++,li2018pointcnn}, are combined with their corresponding point coordinates together as a type of rich point-base feature representation, which are set as input and then fed into deep networks directly for semantic analysis.
Alternatively, geometric properties can be served as auxiliary self-supervision signals, inspired by the recent success of self-supervised learning in visual recognition \cite{doersch2017multi,kolesnikov2019,gan2018geometry,noroozi2018boosting,novotny2018self,fernando2017self,wang2017transitive},
which generate supervision signals from data itself to avoid expensive manual annotations and then learns a proxy loss for network optimization.
Moreover, high-quality local properties preserving finer geometric details can be more accurate in view of more dense sampling of points, which can be provided as privileged supervision signals only available during training.
	
	
Existing geometric learning methods concern on discovering semantic patterns from global shape, which consists of local geometric patterns. 
It remains an open problem whether capturing local geometric patterns have any positive effects on semantic analysis of its global configuration.
This paper is the first attempt to design a novel geometric learning method to explicitly fit local geometric properties in either a self-supervised or a privileged-supervised learning manner as an additional optimization goal to support semantic analysis on point sets.
Fig. \ref{fig:intro} shows the main difference between the proposed geometric learning and conventional supervised classifier.
Specifically, our deep model shares the low-level feature encoding layers and has two branches for semantic analysis (\eg 3D object classification, part/scene segmentation) and geometric properties estimation tasks respectively in a multi-task learning style.
	
The core idea in our work is an auxiliary-supervised learning mechanical, which can boost the performance of general tasks, like classification and segmentation.  Moreover, it is also a multi-task framework, since an additional geometrical loss function is needed. Our method is an orthogonal idea, which can be integrated into different baseline models. Meanwhile, various of different integration methods can also be explored.  Therefore, the novelty of this work is to discover an objective law that can benefit the entire community, that is, adding ‘geometric constraints’ to the 3D deep learning(3DDL) network can improve the performance of different 3DDL tasks. 

The main contributions of this paper are as follows.
\begin{itemize}
\item This work for the first time explores geometric properties of point-based surface, perceiving the underlying local connectivity, as auxiliary supervision signals to improve 3D semantic analysis.
\item A novel geometric self-supervised learning method is proposed to jointly encode feature discriminative for 
semantic analysis on point sets and also well fitting local geometric properties in a multi-task learning manner.
\item Beyond geometric properties via physical computation, high-quality geometric properties as privileged information can further boost performance on semantic analysis.
\end{itemize}
Experimental evaluation on three public benchmarks can demonstrate our motivation to exploit local geometric patterns to improve learning semantic patterns of point clouds, with consistently achieving superior performance to its backbone competitor DGCNN \cite{wang2018dynamic} and other state-of-the-art methods in 3D object classification and part/scene segmentation.

The remainder of this paper is structured as follows. Section II reviews related works with semantic analysis on point cloud. Section III describes the proposed methodology. Section IV demonstrates the detail of our experiments and  discusses the results in this work. The conclusions are presented in Section V. Source codes and pre-trained models can be downloaded at {\color{red}{\url{https://github.com/Necole123/GeoSSL}}}.

	\section{Related Works}
	
\begin{figure*}[thbb]
\begin{center}
\includegraphics[width=0.98\linewidth]{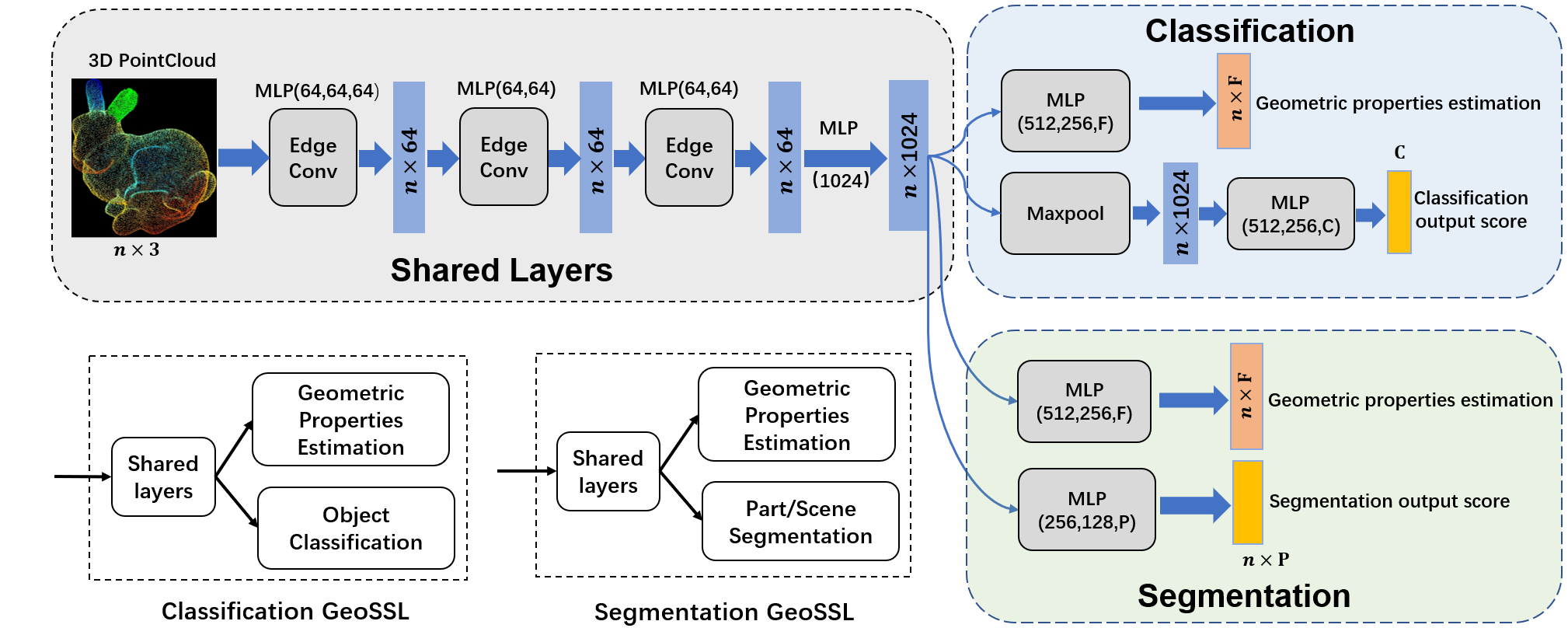}
\caption{
The proposed networks are based on the DGCNN architecture, which aims to estimating local geometric properties and further augmenting semantic analysis of point clouds. 
In Geometric Self-supervised Learning (\textbf{GeoSSL}), the classification network (GeoSSL$_{cls}$) takes $n$ points as input, and shares the first three Edge convolution layers and one MLP layer, which is then divided into two task branches. The top one is the branch to estimate geometric properties, which consists of three fully-connected layers followed with a mean square loss on local geometric properties, while the bottom branch of GeoSSL$_{cls}$ estimates classification scores with the cross-entropy loss. 
The segmentation network (GeoSSL$_{seg}$) shares most of network architecture as GeoSSL$_{cls}$, and the only difference lies in the bottom branch to output the segmentation score on each point. 
Note that, Geometric Privileged Learning (\textbf{GeoPL}) employs the same network, but feeding with high quality of geometric properties as supervision signals in the top branch.} \label{fig:GPNet} 
\end{center} \vspace{-0.1cm}
\end{figure*}

	
\vspace{0.1cm}\noindent\textbf{Semantic analysis of point clouds --} Most traditional features on point cloud are handcrafted towards specific tasks, such as wave kernel signature(WKS) \cite{aubry2011wave}, local reference frame(LRF)\cite{guo2013rotational}, point feature histograms (PFH)\cite{rusu2009fast} and so on. Those point features are often encoded with certain statistical properties or transformed to its 2D counterpart, and are designed to be invariant to certain transformations. In contrast to deep learning based techniques, these hand-crafted point features do not generalize well across different domains. As a pioneer, the PointNet \cite{qi2017pointnet} starts the trend of designing deep networks for operating on irregular point-based surface, with the permutation invariance of points encoded by point-wise manipulation in multi-layer perceptrons (MLPs) and a symmetric function for accumulating features.
Its following work -- Pointnet++ \cite{qi2017pointnet++} hierarchically aggregates multi-scale features to inherently capturing different modes of semantic patterns.
However, both PointNet and PointNet++ only implicitly model semantic concept aware geometric patterns in local regions via deep feature encoding, but miss considering neighborhood information of points to benefit semantic analysis.
Recently, the SO-Net \cite{li2018so} explicitly regularizes spatial correlation across points via $k$-NN search on 2D projection of 3D points during feature encoding, while GeoNet \cite{he2019geonet} implicitly incorporates local connectivity via an autoencoder and a geodesic matching into extra point-wise features for further fusion.
An alternative solution for analyzing point clouds are recently-proposed geometric deep learning methods, such as spectral networks \cite{bruna2013spectral}, which apply convolution operation on graphs representing irregular distributed structure of points.
Its follow-uppers concern on either reducing computational cost by replacing Laplacian eigen decomposition with a polynomial \cite{defferrard2016convolutional,kipf2016semi} and rational \cite{levie2017cayleynets} spectral alternatives, or improving its generalization capabilities \cite{dominguez2018general,zhou2018graph,8822591}.
Recently, DGCNN is proposed by Wang \etal \cite{wang2018dynamic} to discover local geometric manifold of each point by an edge convolution operation on a dynamic $k$-NN graph, which is iteratively updated by the nearest neighbours.
Such a DGCNN model achieves the state-of-the-art performance on semantic analysis, which is thus adopted in our methods as the backbone CNN model.
The key difference between our methods and the DGCNN baseline lies in incorporating an extra branch (as shown in Fig. \ref{fig:GPNet}) to learn local geometric patterns with self-supervision or privileged supervision signals.
Superior performance of our methods can be achieved and illustrated in Tables \ref{table.compareSOTAclass} and \ref{table9} of Sec. \ref{sec.exp}.
	
\vspace{0.1cm}\noindent\textbf{Geometric analysis of point clouds --} Geometric analysis on point clouds aims to obtaining point-wise geometric properties such as the normal and curvature.
A typical solution for obtaining local geometric properties of a point is direct computation based on principle component analysis (PCA) \cite{hoppe1992surface} within a local region, \eg a plane best fitting the point and its $k$-nearest neighbours.
Such a method is simple but sensitive to noises and generation strategies of local regions.
A number of advanced geometric computation techniques \cite{merigot2011voronoi,huang2013edge} are developed to improve robustness against the aforementioned challenges, but remain impractical due to their poor generalization.
On the other hand, geometric shape analysis can be learning-based, \ie learning a regression mapping from point sets to point-wise geometric properties.
A recent deep learning based  PCPNet \cite{guerrero2018pcpnet} performs robustly against noises and shape variation under a wide variety of settings, with sufficiently large-scale training data.
Our goal of this paper is to directly mine local geometric patterns to additionally support semantic analysis on point-based shape via an auxiliary supervised mapping onto geometric properties.
In our proposed multi-task network, more robust estimation on geometric properties can be achieved than fixed backbone baseline (See Fig. \ref{fig:compare} and \ref{fig:normal_error}), with also improving classification accuracy for semantic analysis (See Table \ref{table.compareSOTAclass}). 
	
\vspace{0.1cm}\noindent\textbf{Deep self-supervised learning --} Deep learning has gained significant successes in visual recognition \cite{krizhevsky2012imagenet,zhang2019part,wu2018deep,jia2019orthogonal} and semantic shape analysis \cite{qi2017pointnet,qi2017pointnet++,wang2018dynamic, wu2016learning,li2018so,yang2018foldingnet} but heavily hinges on large-scale labelled training samples.
Data augmentation becomes a simple yet effective pre-processing step to alleviate the demand for sufficient data to fit network parameters, especially for the larger network capacity than size of training samples.
For avoiding label acquisition for some supervision-starved tasks and using vast numbers of unlabelled data, self-supervised learning is considered as a powerful alternative to relax the impractical requirement about large-scale labelled data available, via generating supervision labels from data itself.
In other words, the self-supervised learning paradigm is typically formulated into a \textit{pretext} learning task, such as motion segmentation in videos \cite{pathak2017learning}, and relative positions \cite{doersch2015unsupervised}, exemplars \cite{7312476} in the image domain. 
In light of this, the \textit{target} task can be solved through transferring knowledge from self-supervised learning on a proxy loss.
Inspired by the concept in self-supervised learning, this paper for the first time develops a novel geometric self-supervised learning (GeoSSL) to exploit local geometric patterns discovered by self-supervised learning to improve semantic analysis of point clouds.
With local geometric regularization on deep feature encoding for semantic analysis, experiment results of the proposed GeoSSL can beat its direct competitor -- DGCNN (the backbone net) as well as other comparative methods (see Table \ref{table.compareSOTAclass}).
	

\vspace{0.1cm}\noindent\textbf{Learning with privileged information --}
Information only available during training is referred to privileged information, which has been exploited in
classification \cite{vapnik2009new,LapHeiSch:2014}, regression \cite{yang2013privileged} and ranking \cite{sharmanska2013learning}.
For image based semantic analysis, text \cite{sharmanska2013learning}, attributes \cite{sharmanska2013learning}, bounding boxes \cite{sharmanska2013learning}, head pose \cite{yang2013privileged}, and gender \cite{yang2013privileged} have been exploited as privileged information to boost performance, but this paper is the first work, to the best of our knowledge, in geometric learning with high-quality properties from more densely sampled points as privileged information.
Similar to the aforementioned GeoSSL method, our geometric privileged learning (GeoPL) employs the identical multi-task network structure, and the only difference between GeoSSL and GeoPL lies in the quality of geometric properties to discover local patterns of 3D geometry to support semantic classification and segmentation. Experimental verification in this paper demonstrates that our model with privileged geometric properties performs better than the state-of-the-art methods in Table \ref{table.compareSOTAclass} as well as its self-supervised variant.
	
	
\section{Methodology}\label{sec.methodology}

\subsection{Supervised Semantic Learning}\label{sec.ssl}
	
	
Existing deep algorithms on point clouds focus on analysing semantic patterns of 3D geometry, in view of only semantic labels available in 3D object classification \cite{wang2018dynamic} or part segmentation \cite{li2018pointcnn}.
Given a pair of 3D observation in the representation of a point cloud $\bm{P}$ and its semantic label $y$, the typical network architecture of supervised semantic learning frameworks such as PointNet \cite{qi2017pointnet}, PointNet++ \cite{qi2017pointnet++}, PointCNN \cite{li2018pointcnn} and DGCNN \cite{wang2018dynamic} consists of several feature encoding layers (\eg convolutional layers, MLP layers or a hybrid of both).
Take the DGCNN \cite{wang2018dynamic} (the backbone network of the proposed GeoSSL and GeoPL) as an example, which is shown in gray box of Figure \ref{fig:GPNet}.
The	DGCNN introduces edge convolution operation on a directed graph representation for local connectivity of points. 
In details, a directed $k$-Nearest Neighbour ($k$NN) graph $\mathcal{G} = (\mathcal{V}, \mathcal{E})$ models correlation across closest vertices, where $\mathcal{V} = {1,2,\ldots,k}$ and $\mathcal{E} \subseteq  \mathcal{V} \times \mathcal{V}$ denotes its vertices and edges. 
A parametric mapping function on edges $f_{\mathcal{\theta}}(\mathcal{V}_i,\mathcal{V}_j) = f_{\mathcal{\theta}}(\mathcal{V}_i,\mathcal{V}_j -\mathcal{V}_i)$ is adopted for capturing global and local shape patterns, where $\mathcal{\theta}$ is the parameters to be optimized in each edge convolution layer. 
In this sense, the output of edge convolution on the $k$-NN graph on each vertex is calculated by aggregating $k$ edge features, which is thus invariant to the total size of points in the set.
	
Shared parts of the DGCNN is made up of three MLP based edge convolution blocks and a fully-connected layer to encode each point into a 1024-dimensional feature, and task-specific layers for object classification and part segmentation respectively.
On one hand, another multi-layer perception, the output dimension of hidden layers in each MLP based decoder fixed to \{512, 256, $C$\}, where $C$ denotes the size of object classes, is added to the shared parts of the DGCNN for semantic object classification.
On the other hand, the shared parts of the DGCNN is followed by a multi-layer perception with \{256, 128, $P$\}, where $P$ denotes the size of object part classes in part segmentation.
However, such a model cannot provide supervision signals to incorporate local geometric structural information,
which encourages us to design a novel network for improving semantic analysis by learning primitive geometric properties of points in their local neighbourhood.
	

\subsection{Generation of Local Geometric Properties}\label{SecSelfSupervision}
	
Given a point set $\bm{P}$, point-wise geometric properties can be either measured or calculated directly. 
A typical solution of generating $i$th point's normal is first to find out its $k$-nearest neighbors $G=\{ \bm{p}_{1}^{'},\bm{p}_{2}^{'}, \dots, \bm{p}_{k}^{'} \} $ and then calculate the covariance matrix $\bm{C}$ as
\begin{equation}
\bm{C} = \sum_G \bm{r}\bm{r}^T,
\end{equation}
where $\bm{p}$ denotes points in the cloud and $\bm{r} = \bm{p}_{i} - \bm{p}_{j}^{'} , ~j= \{1,2,\ldots,k\}$.
Eigenvectors $\bm{e}$ and eigenvalues $\lambda$ of $\bm{C}$ can be obtained via spectral decomposition \cite{bae2008method}.
The eigenvector  corresponding to the minimal eigenvalue defines the estimated surface normal $\bm{n}_{p_{i}}$ of point $\bm{p}_i$, as defined in \cite{Tatarchenko2018CVPR}. 
Similarly, the second-order geometric property -- curvature can also be calculated based on eigen decomposition on covariance matrix $\bm{C}$ \cite{bae2008method}. 
Particularly, the ratio of the minimal eigenvalue and the sum of all the eigenvalues can be used to estimate the change of geometric curvature.
In mathematics, for $i$-th point $\bm{p}_i$, the change of curvature $u_i$ can be approximated as the following
\begin{equation}
u_{p_{i}} = \frac{\lambda_\text{min}}{\sum \lambda},
\end{equation}
where $\lambda_\text{min}$ denotes the minimal eigenvalue of $\bm{C}$.
Additionally, for $i$-th point $\bm{p}_i$, the curvature $u_{p_{i}}$ can also be computed by the normal vectors of that point and its neighbors as
\begin{equation}
u_{p_{i}}=\frac{1}{k}\sum_{j=1}^{k}\left \|\bm{n}_{p_{i}}-\bm{n}_{neighour(j,p_{i})} \right \|.
\end{equation}
Although geometric properties can be directly computed from point clouds, they can also be estimated via supervised regression learning algorithms \cite{guerrero2018pcpnet,ben2018nesti}.
	
Normal and curvature approximating local geometric patterns of the shape are vital in semantic analysis, which encourages a number of work \cite{qi2017pointnet,qi2017pointnet++} to combine such point-wise geometric properties with their corresponding coordinates, which are then fed into a supervised learning model as feature input.
However, very few works consider normal and curvature of points as auxiliary supervision signals to improve analyzing semantic patterns owing to feature encoding local manifold structure and superior robustness against noisy point sets, especially when the model is trained on clean data. 	
Beyond point-wise normal and curvature by computational self-generation from point clouds, more accurate and high quality geometric properties can be provided as privileged information available only during training, \eg via physical computation from more dense points.

\subsection{Multi-Task Geometric Learning}
	

In view of lack of local connectivity across order-less points, our motivation is to design an auxiliary task (regression learning with geometric properties) to explicitly incorporate local neighborhood information underlying surface manifolds.
To this end, we propose a multi-task geometric learning network to simultaneously learn semantic and geometric patterns for 3D object classification and part segmentation, whose pipelines are visualized in Fig. \ref{fig:GPNet}.
Given input and output pairs for an ordinary supervised learning network, \ie a point cloud $P$ and its semantic class labels $y$, geometric properties $\bm{g}$ can be generated by physical computation in Sec. \ref{SecSelfSupervision} as extra self-supervision signals or provided as privileged information extracted from high quality point clouds, \ie Geometric Self-Supervised Learning (GeoSSL) and Geometric Privileged Learning (GeoPL) respectively.
It is noted that, regardless of qualities of auxiliary labels, the proposed networks have an identical network structure for classification or segmentation. 
Training pairs for our multi-task geometric learning network are thus $\{P, \bm{g}, y\}^{i=1,2,\ldots,N} $, where $\bm{g} = (\bm{n}, u)_i^{i=1,2,\ldots,N} \in \mathbb{R}^4$ denotes point-wise geometric properties and $N$ is the size of training samples.
	
Based on the backbone DGCNN depicted in Sec. \ref{sec.ssl}, the proposed geometric learning consists of the shared layers and the application-specific block (blue or green boxes in Fig. \ref{fig:GPNet}), which shares the first three Edge convolution blocks followed by one MLP layer and is divided into two task-specific branches. 
The top branch is an auxiliary task to regress point-wise local geometric properties, while the bottom one is the original tasks of semantic analysis (\ie classification, part/scene segmentation).
To jointly optimizing both branches, we introduce a combinational loss function as the following, which utilizes the mean square loss $L_{reg}(\bm{g},\hat{Y}_{reg})$ to control the quality of normal/curvature estimation in geometric learning branch and the cross-entropy loss $L_{task}(y,\hat{Y}_{task})$  for task-specific semantic analysis on point sets as:
\begin{equation}
arg \underset{\theta_{s},\theta_{task},\theta_{reg}}{min} L_{task}(y,\hat{Y}_{task})+\lambda L_{reg}(\bm{g},\hat{Y}_{reg}), \label{loss}
\end{equation}
where $\hat{Y}_{task}$ and $\hat{Y}_{reg}$ denote output of two branches in the proposed model, and $\theta = \{\theta_{s},\theta_{task},\theta_{reg}\}$ are weighting parameters of the proposed geometric learning model. 
$\theta_{s}$ denotes shared weights in the lower shared layers, and $\{\theta_{task},\theta_{reg}\}$ are weights for the  classification/segmentation  and the geometry regression branch, respectively. 
$\lambda$ is a trade-off parameter between two loss terms.

The key merit of the aforementioned cost function lies in that it brings additional object function to discover geometric patterns missing by existing supervised point cloud classifiers trained by semantic labels only.
During training, we adopt the mean square loss for $L_{reg}$ and the cross-entropy loss for $L_{task}$. 
It is noted that the regression loss is not limited to the mean square, and we select it owing to its solid performance on estimation of geometric properties.
Specifically, we have explored the Euclidean distance and Cosine similarity for the oriented normal vector, the unoriented normal Euclidean distance and RMS angle difference between the estimated normal and ground truth normal in our experiments.
Without the loss of generality, we also employ the mean square loss for supervising geometric curvature. As a result, with both normal and curvature, the loss function $L_{reg}$ can be written as
\begin{equation}
L_{reg}=\frac{1}{m}\sum_{i=1}^m\|\bm{n}_i-\hat{\bm{n}}_i\|^2 + \frac{1}{m}\sum_{i=1}^m(u_i-\hat{u}_i)^2
\end{equation}
where $\bm{n}_i$ and $\hat{\bm{n}}_i$ denote the ground truth normal (self-generated in GeoSSL or privileged provided in GeoPL) and the predicted normal, and $u_i$ and $\hat{u}_i$ denote the ground truth curvature and the predicted curvature.

	
	
	
\begin{figure}[t]
\begin{center}
\includegraphics[width=0.9\linewidth]{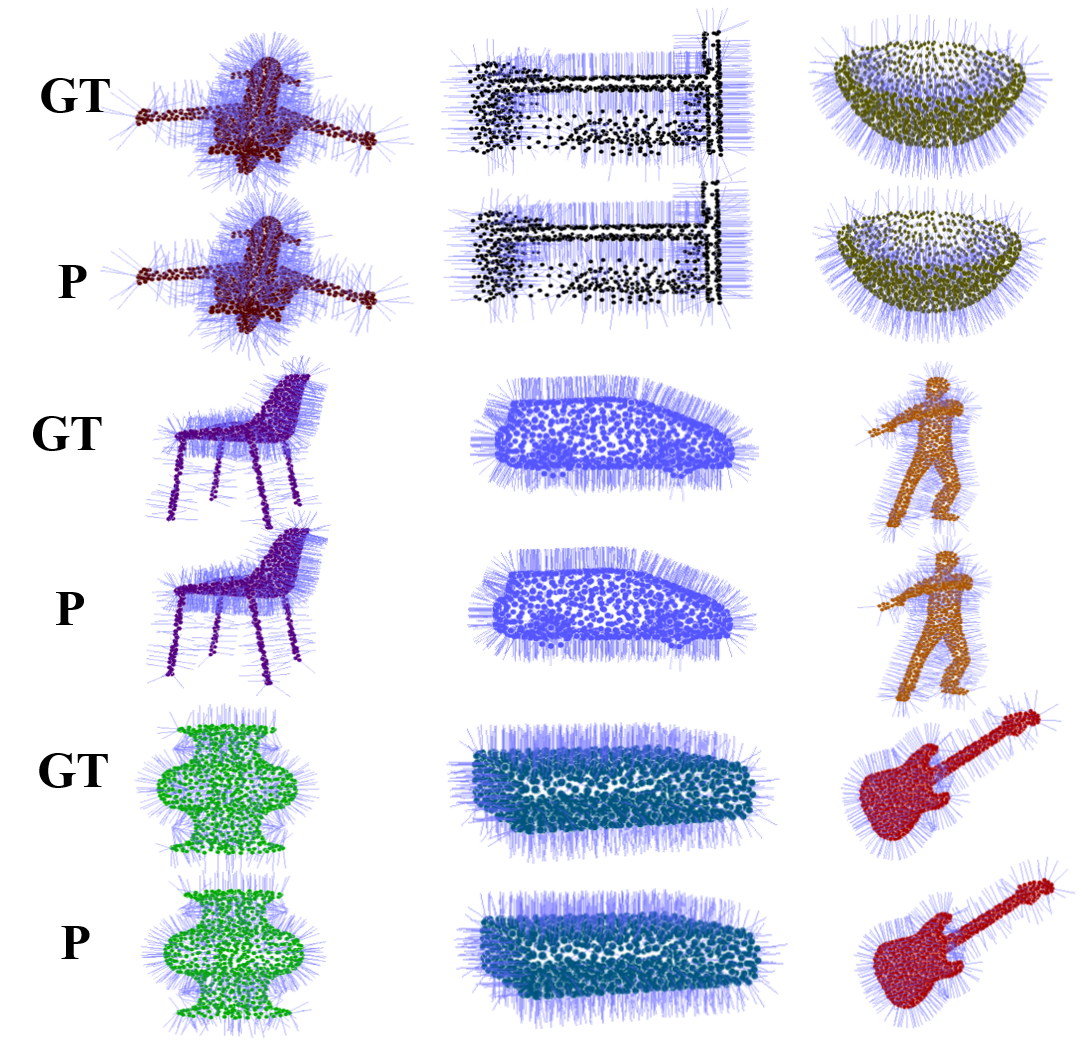}
\caption{Visualization of predicted normal with GeoSSL$_\text{dgcnn}$.} \label{fig:normal}
\end{center}
\end{figure}
	
\begin{figure}[t]
\begin{center}
\includegraphics[width=0.9\linewidth]{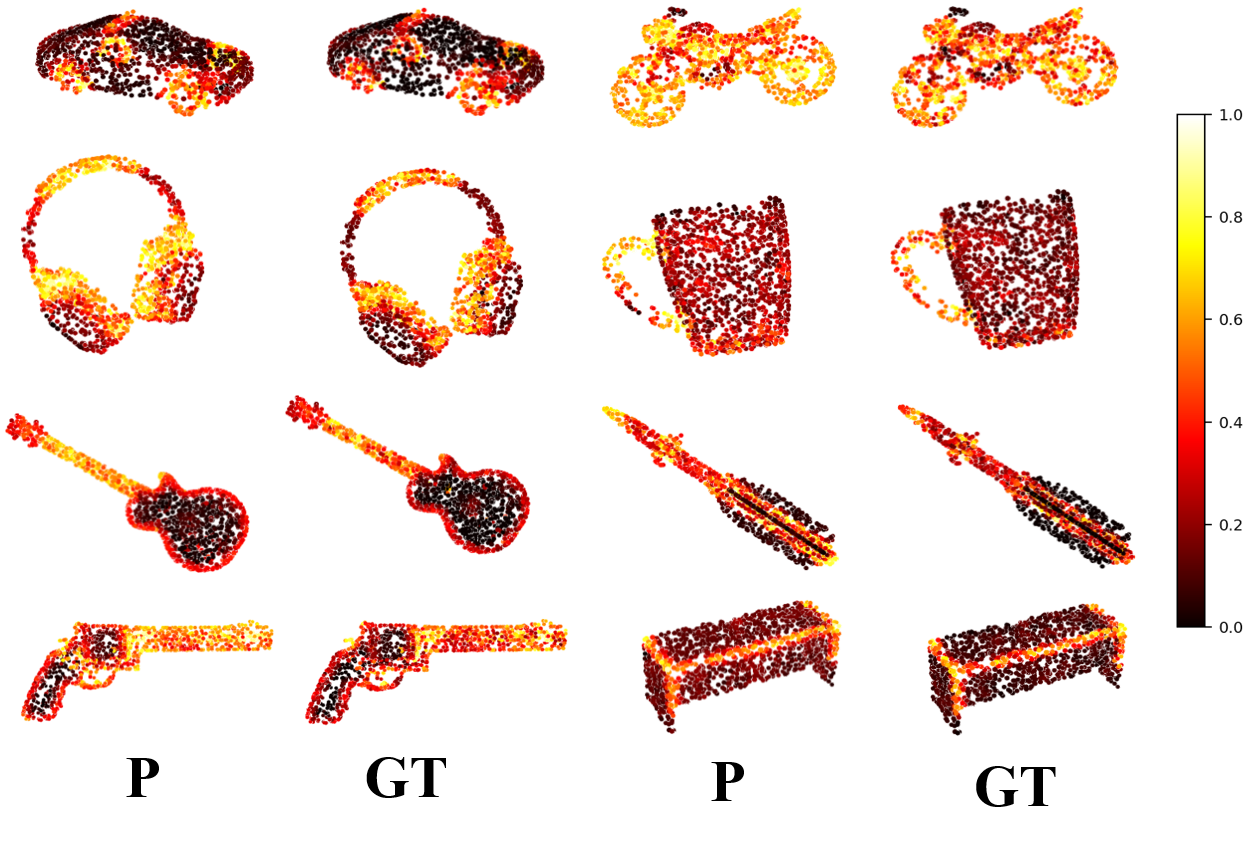}
\caption{Visualization of predicted curvatures with GeoSSL$_\text{dgcnn}$ with a color bar on the right hand side. The darker, the smaller value of curvature.} \label{fig:cur}
\end{center}
\end{figure}

\section{Experiments}\label{sec.exp}
	
We evaluate the proposed geometric learning algorithms (\ie GeoSSL and GeoPL) introduced in Sec. \ref{sec.methodology} on three popular semantic analysis tasks, \ie 3D object classification, part segmentation and scene segmentation.
	
	
\vspace{0.1cm}\noindent\textbf{Datasets and Settings --} Evaluation on 3D object classification was conducted on the commonly used ModelNet40 benchmark \cite{wu20153d}, which contains 12,311 CAD models belonging to 40 pre-defined categories.  
In our experiments, we split the dataset into two parts, \ie 9,843 for training and 2,468 for testing.  We followed the same experimental settings as in \cite{qi2017pointnet,wang2018dynamic}. 
Specifically, 1024 points are sampled from mesh faces by farthest point sampling, and are normalized into a unit sphere. 
We evaluated our model architectures for part segmentation on the ShapeNet part dataset \cite{yi2016scalable}, containing 16,880 3D shapes from 16 object categories, annotated with 50 parts in total. 
We followed the data split as \cite{li2018pointcnn}, \ie 14006 for training and 2874 for testing. Part category labels are assigned to each point in the point cloud, which consists of 2048 points uniformly sampled from mesh surfaces of training samples. It is worth mentioning here that we assume that each object contains less than six parts.
S3DIS \cite{armeni20163d} dataset is adopted on evaluation of our method for scene segmentation. Unlike the samples in the ModelNet40 and ShapeNet, which are made by 3D modeling tools, the S3DIS samples are collected from real scans of indoor environments. In details, this dataset contains 3D scans from Matterport scanners in 6 areas within 271 rooms. Each point in the scan is annotated with one semantic label from 13 categories.

\vspace{0.1cm}\noindent\textbf{Performance Metrics --} For the classification task, we use mean accuracy (mA) as our evaluation metric widely adopted in recent work \cite{qi2017pointnet,qi2017pointnet++,wang2018dynamic}.

{\begin{equation}
mA = \frac{1}{K}\sum\limits_{k=1}^{K} \frac{\sum\limits_{i=1}^{N}I_k(\hat y_i)}{\sum\limits_{i=1}^{N}I_k( y_i)}
\end{equation}
where $K$ is the number of categories in test set, $N$ is the number of testing samples. $I_k(y)$ is the indicator function, when $y=k$, $I_k(y) = 1$, otherwise $I_k(y) = 0$. $y_i$ and $\hat y_i$ are the true label and predicted label of the $i$th sample  respectively.

In the part segmentation task, Intersection-over-Union (IoU) is used to evaluate our model and other comparative methods, following the same evaluation protocol as the DGCNN \cite{wang2018dynamic}, the IoU of a shape is obtained by averaging the IoUs of different parts involving in that shape, while the mean IoU (mIoU) is calculated by averaging the IoUs of all the testing samples.	
\begin{equation}
IOU_c^i = \frac{|P_{\hat j =c}^i \cap P_{j=c}^i|}{|P_{\hat j =c}^i \cup P_{j=c}^i|} 
\end{equation}
\begin{equation}
IOU^i = \frac{1}{C}\sum\limits_{c=1}^{C} IOU_c^i
\end{equation}
\begin{equation}
mIOU = \frac{1}{N}\sum\limits_{i=1}^{N} IOU^i
\end{equation} 
\noindent where $C$ is the number of categories need to be split in each sample. $N$ is the number of testing samples. $P_{j=c}^i$ indicates the subset of category $c$ in point cloud $P^i$. $P_{\hat j =c}^i$ represents the subset  that be correctly divided into category $c$ in $P^i$, then $IOU_c^i$ is the IOU of the $c$th  category in $P^i$, and $IOU^i$ is the IOU of point cloud $P^i$.

In the scene segmentation task, mean Intersection-over-Union (mIoU) and overall accuracy(OA) are utilized for evaluating our method as follows
\begin{equation}
OA = \frac{|\{p|p\in \mathcal{P}\cap \hat y_p=y_p\}|}{|\mathcal{P}|}
\end{equation}
\noindent where $|\mathcal{P}|$ is the numbers of point in evaluate dataset $\mathcal{P}$, and $|\{p|p\in \mathcal{P}\cap \hat y_p=y_p\}|$ is the number of correctly segmented points.

\vspace{0.1cm}\noindent\textbf{Implementation Details --} To efficiently use the geometric cues, we pre-train independently the  shared layers and geometric leaning branch (the top branches in Fig. \ref{fig:GPNet}) with generated geometry properties on the ShapeNetCore dataset, which is similar to the DGCNN architecture for part segmentation with the only change lying in the last layer to output 4 continuous values. 
Model parameters learned by such a network are then used to initialize the shared layers both in GeoSSL$_\text{cls}$ and GeoSSL$_\text{seg}$.
The learning rates of the GeoSSL$_{cls}$ and GeoSSL$_{seg}$  are set as 0.01 and 0.001 respectively, and are decreased with an exponential function by every 20 epochs. 
The overall training epochs in our experiments are 200. 
	

\subsection{Comparison with State-of-the-Art}
	
	
\vspace{0.1cm}\noindent\textbf{3D object classification --} Comparative evaluation in 3D object classification on the ModelNet40 are shown in Table \ref{table.compareSOTAclass}. We can see that our  GeoSSL$_{dgcnn}$ achieves superior performance to its direct competitor DGCNN \cite{wang2018dynamic} as well as other state-of-the-art methods. In light of the identical input and output as well as the backbone CNN model, performance gain can only be explained by auxiliary incorporation of local geometric properties into the DGCNN. 	
We also evaluate our geometric privileged learning (GeoPL) for classification on the ModelNet40 with privileged geometric properties only available during training, whose normal and curvature are generated from more dense point-based surface and thus more accurate than those directly computed from sparse points. 
For example, we can generate privileged normal and curvature from a dense point cloud consisting of 10000 points used in our experiment compared to ordinary one with 1024 points. 
Experiment results in Table \ref{table.compareSOTAclass} show significantly better performance than other comparative algorithms given accurate geometric properties, which further verifies the effectiveness of our concept on improve semantic analysis via exploiting local geometric priors.

\begin{table}[t]
\caption{Comparisons of classification accuracy on the ModelNet40. Note that, the DGCNN and DGCNN+ here denote the DGCNN in \cite{wang2018dynamic} without and with the spatial transformer respectively. Our GeoSSL and GeoPL adopt the former as their backbone. }\label{table.compareSOTAclass}
\centering
\begin{centering}
\setlength{\tabcolsep}{3mm}
\begin{tabular}{l|c|c}
\hline
Methods & Mean  & Overall \\
& Class Accuracy & Accuracy \\
\hline
VoxNet \cite{maturana2015voxnet} & 83.0 & 85.9\\
PointNet \cite{qi2017pointnet} & 86.0 & 89.2 \\
PointNet++ \cite{qi2017pointnet++} & - & 90.7\\
SO-NET \cite{li2018so} & 87.3 & 90.9\\
PointCNN \cite{li2018pointcnn} & - & 92.2 \\
\hline
DGCNN \cite{wang2018dynamic} & 88.2 & 91.2 \\
DGCNN+ \cite{wang2018dynamic} & 90.2 & 92.2 \\
GeoSSL$_\text{dgcnn}$ (ours) & {90.3} & {92.9} \\
GeoPL$_\text{dgcnn}$(ours) & \textbf{90.8} & \textbf{93.5} \\
\hline
\end{tabular}
\end{centering}
\end{table}
	
\vspace{0.1cm}\noindent\textbf{3D Part segmentation --} The part segmentation network is evaluated on the ShapeNet Part benchmark, whose results on Intersection-over-Union (IoU) are illustrated in Table \ref{table.segSOTA}. 
Evidently, regardless of the network structure, \eg PointNet++ \cite{qi2017pointnet++}, PointCNN \cite{li2018pointcnn} or DGCNN \cite{wang2018dynamic}, the proposed GeoSSL can consistently perform better than the backbone competitors. Specifically, PointNet++ \cite{qi2017pointnet++} achieves the better performance compared to our GeoSSL but demands high quality point-wise geometric properties as input, which can be impractical for accurate point-wise normals available in the real world. We re-implement PointNet++, PointCNN and DGCNN by following the settings in original works, but slightly change the input or network architectures, whose results are reported\footnote{Our Implementation is slightly worse than the reported results in the original works.} and noted as the backbone in each block of Table \ref{table.segSOTA}. It is noted that the input of original PointNet++ is coordinates combined with normal, while our backbone PointNet++ only utilizes coordinates as input, which needs to be consistent with our proposed GeoSSL. For backbone DGCNN, we slightly change the network architecture by removing the spatial transformer module in the original DGCNN. Because we need to calculate the geometric properties by the neighbor information of each point. While spatial transformer module may destroy the local surface structure. Therefore, Our GeoSSL and GeoPL adopt the DGCNN without spatial transformer as their backbone.

In view of the identical network structure to capture semantic properties in the GeoSSL and its backbone baselines, performance gain can only be explained by exploiting local fine-detailed geometries of objects, which can demonstrate our motivation again. 
More results about part segmentation results are illustrated in Fig. \ref{fig:seg_v} and Fig.\ref{fig:compared_v}, from which we can see that the segmentation results of our method are very close to the ground truth and always better than its backbone.

	
\begin{table}[t]
\caption{Comparisons of part segmentation results on the ShapeNet part dataset with Mean IoU (\%). }\label{table.segSOTA}
\vspace{0.1cm}	
\centering
\begin{centering}
\setlength{\tabcolsep}{8mm}
\begin{tabular}{l|c}
\hline
Methods & Mean IoU\\
\hline
PointNet \cite{qi2017pointnet}&83.7 \\
SO-NET \cite{li2018so} &84.9\\
\hline
PointNet++ \cite{qi2017pointnet++} &85.1 \\
PointNet++ (backbone) &84.3 \\
GeoSSL$_\text{pointnet++}$ (ours) &84.8 \\
\hline
PointCNN \cite{li2018pointcnn}&86.1\\
PointCNN (backbone) &85.3\\
GeoSSL$_\text{pointcnn}$ (ours) &85.6\\
\hline
DGCNN+ \cite{wang2018dynamic} &85.1\\
DGCNN (backbone) &84.5 \\
GeoSSL$_\text{dgcnn}$ (ours) &85.7 \\
\hline
\end{tabular}
\end{centering}  
\end{table}
		
\begin{figure}[t]
\begin{center}
\includegraphics[width=1\linewidth]{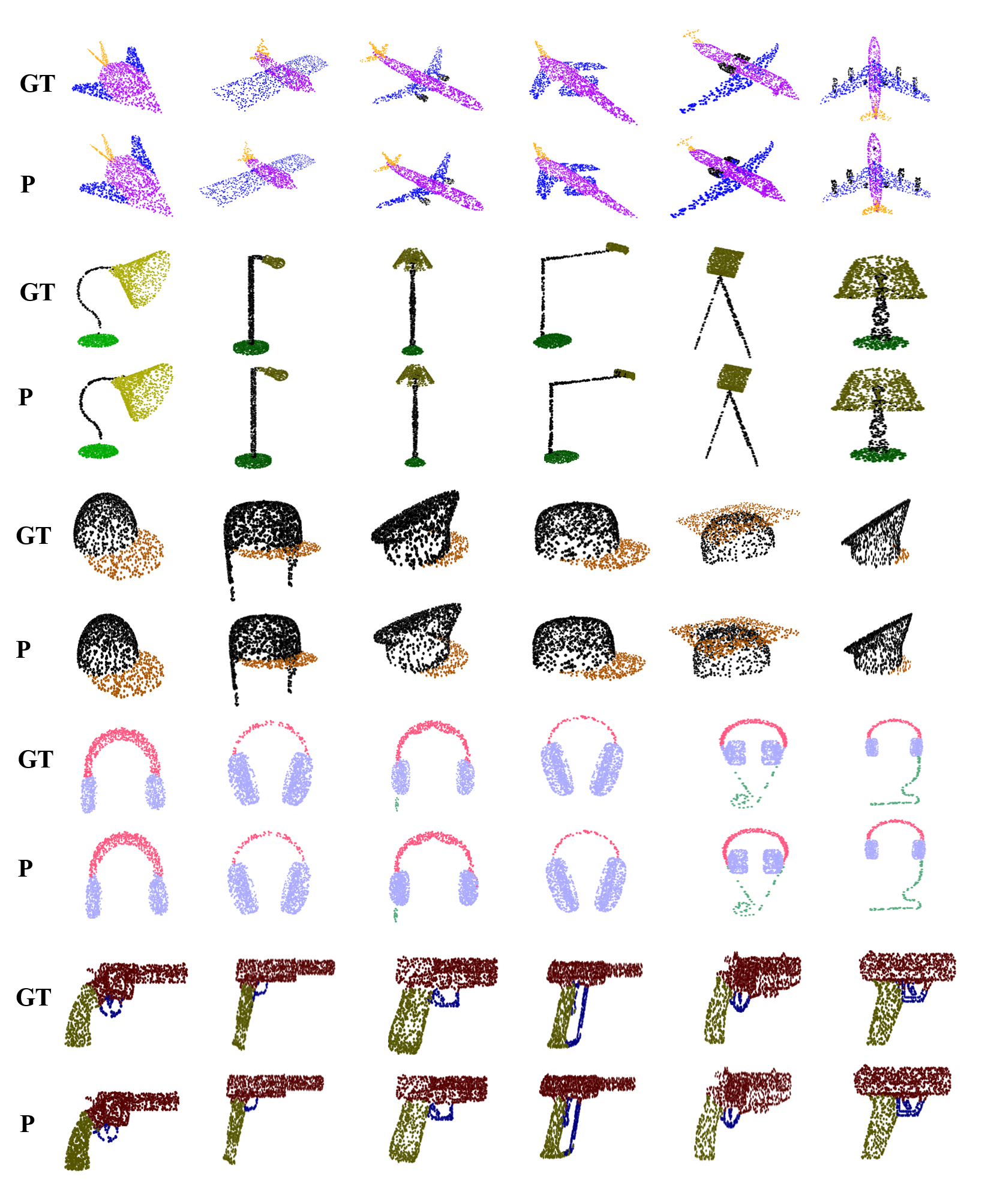}
\caption{Visualization of part segmentation results with GeoSSL$_\text{dgcnn}$, where GT denotes ground truth label, and P means predicted result.} \label{fig:seg_v}
\end{center}
\end{figure}	
				
\begin{figure}[t]
	\begin{center}
		\includegraphics[width=0.98\linewidth]{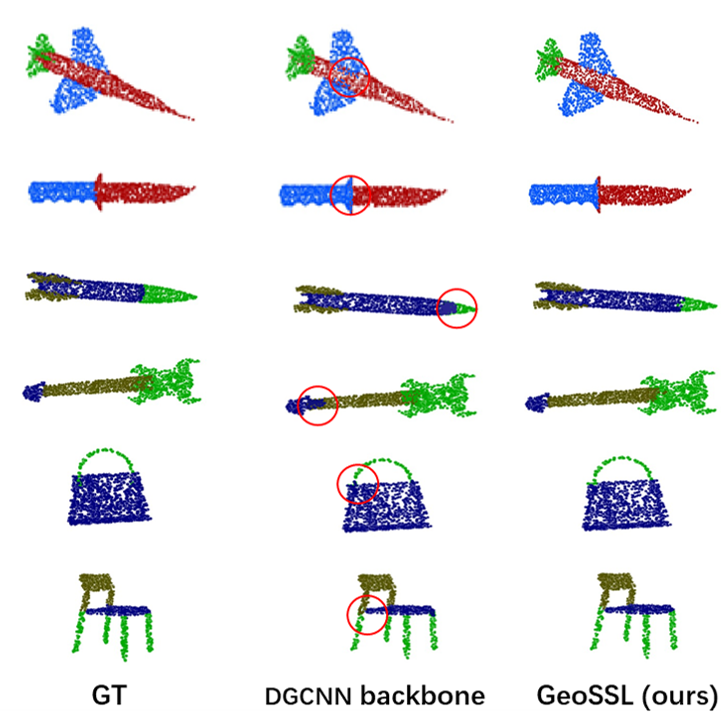}
		\caption{{Comparisons with segmentation results by the proposed GeoSSL$_\text{dgcnn}$ (ours) and backbone DGCNN.}} \label{fig:compared_v}
	\end{center}
\end{figure} 	
	
\vspace{0.1cm}\noindent\textbf{Indoor Scene Segmentation --} We also apply our GeoSSL to the semantic scene segmentation task, which replaces object part labels in part segmentation by semantic object classes in the scene.
We conduct experiments on the S3DIS\cite{armeni20163d}, which is collected from real scans of indoor environments. For a fair comparison, we follow the same setting as the DGCNN, where each room is sliced into 1 $\times$ 1 square-meters block, and 4096 points are sampled for each block. 
Based on the sampled points, we then calculate point-wise geometric properties (\ie normal, curvature) using the method in Sec. \ref{SecSelfSupervision}.
Finally, we use the  6-fold cross validation over the 6 areas, and report the mean of evaluation results. 
We compare the proposed method with the state-of-the-art methods on the S3DIS, whose results are shown in Table \ref{seg_comparison}. 
We can conclude that our method consistently achieves superior segmentation performance to its direct competitor DGCNN \cite{wang2018dynamic}, yet outperforms most of state-of-the-art methods except for PointCNN \cite{li2018pointcnn} and SPGraph \cite{landrieu2017large}.  
Note that, the concept of our method is generic, which can be applied to other specific backbone CNN models, which achieves state-of-the-art scene segmentation performance, such as PointCNN \cite{li2018pointcnn} and SPGraph \cite{landrieu2017large}. 
		
\begin{table}[t]
\caption{Segmentation comparisons on S3DIS in mean IoU (mIoU, \%) and overall accuracy (OA, \%). }\label{seg_comparison}\centering
\begin{centering} 
\begin{tabular}{l|c|c}
\hline
\multirow{2}{1.5cm}{\centering Methods} & Mean & Overall \\
& IoU & Accuracy \\
\hline
PointNet(baseline) \cite{qi2017pointnet} & 20.1 & 53.2\\
PointNet \cite{qi2017pointnet} & 47.6 & 78.5 \\
PointCNN \cite{li2018pointcnn} & 65.4 & 88.1 \\
G+RCU \cite{engelmann2017exploring} & 49.7 & 81.1\\
SGPN \cite{wang2018sgpn} & 50.4 & -\\
RSNet \cite{huang2018recurrent} & 56.5 & -\\
SPGraph \cite{landrieu2017large} & 62.1 & 85.5\\
\hline
DGCNN+ \cite{wang2018dynamic} & 56.1 & 84.1 \\
DGCNN$_\text{(our baseline)}$ &54.5 & 83.6 \\
\textbf{GeoSSL$_\text{dgcnn}$} & 59.1 & 86.3 \\
\hline
\end{tabular}
\end{centering}
\end{table}

\subsection{More Results and Discussions}\label{sec.ablation}
	
\begin{table}[h]\centering
\caption{Points ($P$) with vs. without geometric properties including normal ($\bm{n}$) and curvature ($\bm{u}$) on  mean classification accuracy (\%). }\label{table7}
\centering
\begin{centering}
\setlength{\tabcolsep}{4mm}
\begin{tabular}{l|c|c}
\hline
Methods & DGCNN\cite{wang2018dynamic} & GeoSSL$_\text{dgcnn}$  \\
\hline
$P$  & 91.2 & 92.2 \\
\hline
$P$ + $\bm{n}$  & 91.7 & 92.5 \\
\hline
$P$ + $\bm{u}$ & 91.4 & 92.3 \\
\hline
$P$ + $\bm{n}$ + $\bm{u}$ & 91.9 &	\textbf{92.9} \\
\hline			
\end{tabular}
\end{centering}
\end{table}	

\vspace{0.1cm}\noindent\textbf{Ablation studies on geometric properties --} Evaluation on combination of different geometric properties is shown in Table \ref{table7}. In DGCNN \cite{wang2018dynamic}, geometric properties are concatenated as additional feature input, while our GeoSSL exploits them as self-supervision signals of an auxiliary task. 
We observe that all methods with geometric properties either as input feature or as self-supervision signals can boost classification performance, which demonstrates our motivation to employ local geometric properties can reveal rich local geometries of 3D semantic classes. 
Moreover, geometric properties as self-supervision signals (in the right column) can consistently perform better than that as feature (in the middle column). 
The main reason is that our GeoSSL takes the form of multi-task learning, where self-supervision serves an auxiliary task to regularize learning of the main, supervised task. 
This is different from some alternatives, \eg pre-training based self-supervision methods, where features are learned via self-supervision alone, and are subsequently used for supervised tasks. 
Given large capacities of deep networks, GeoSSL regularizes feature learning (via self-supervised prediction learning of local geometric properties), reduces their potentials of over-fitting, and thus improves generalization of the learned features for the supervised tasks.
Moreover, the combination with normal and curvature can be preferred as self-supervision signals in view of exploiting both first and second order geometric smoothness in point sets. 	
		
\begin{figure}[t]
\begin{center}
\includegraphics[width=0.85\linewidth]{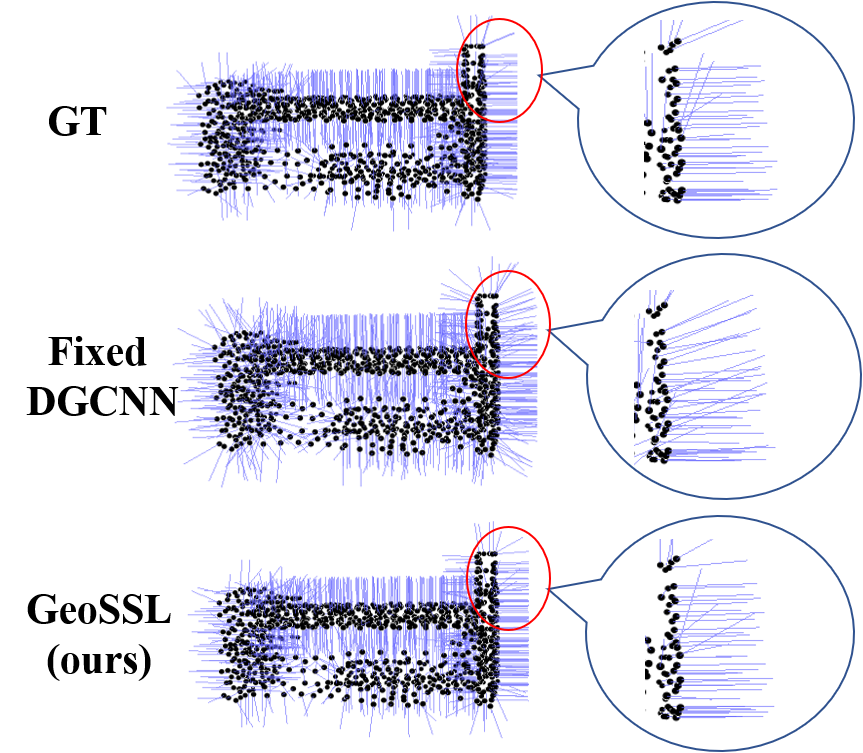}
\caption{Comparisons with learned geometric properties by the proposed GeoSSL$_\text{dgcnn}$ (ours) and DGCNN$_{fixed}$.} \label{fig:compare}
\end{center}
\end{figure}

\begin{table}[h] \centering
\caption{Comparison of the Cosine Similarity for normal estimation with involved methods} \label{table8}
\vspace{0.1cm}
\setlength{\tabcolsep}{7mm}
\begin{tabular}{l|c}
\hline
Methods  & Cosine Similarity \\
\hline
DGCNN \cite{wang2018dynamic} & 0.99\\
DGCNN$_{fixed}$ & 0.97 \\
\hline
GeoSSL$_{dgcnn}$ & 0.99 \\
\hline
\end{tabular}
\end{table} 
		
\vspace{0.1cm}\noindent\textbf{Effects of learning geometric patterns in typical supervised semantic learning --} We are interested in whether the learned feature in supervised semantic learning on  point clouds can be used to estimate geometric properties.
As a result, we conduct an experiment for normal estimation to compare the following models:
the first setting is to train the DGCNN for normal estimation from scratch, denoted as DGCNN in Table \ref{table8}; the second setting is another DGCNN, whose network parameters of lower layers are shared by the DGCNN  pre-trained on the ModelNet40 for classification, which are then fixed during training with tuning the other parameters in higher layers (we denote it as Fixed DGCNN (DGCNN$_{fixed}$).  
The results are illustrated in Table \ref{table8} for a comparative purpose on the Cosine Similarity metric, which reveals an angle difference between the predict normal and the ground truth normal, \ie the larger its value, the better. 
We also illustrate qualitative difference between our method and DGCNN$_{fixed}$ in Fig. \ref{fig:compare}, which shows that the proposed method can predict more accurate normal than its competitor.
Quantitative comparisons with normal estimation errors can be found in Fig. \ref{fig:normal_error}. 
Both Table \ref{table8}, Fig. \ref{fig:compare} and \ref{fig:normal_error} show that the DGCNN$_{fixed}$ gain the worse performance in comparison with the DGCNN and GeoSSL$_\text{dgcnn}$. 
It implies that existing  point cloud analysis methods with only semantic supervision labels pay less attention on whether the networks can learn local geometric patterns. 
Our method with geometric self-supervised learning can benefit each other task simultaneously, which captures local geometric patterns to further augment semantic recognition tasks.

\begin{figure}[t]
\begin{center}
\includegraphics[width=0.9\linewidth]{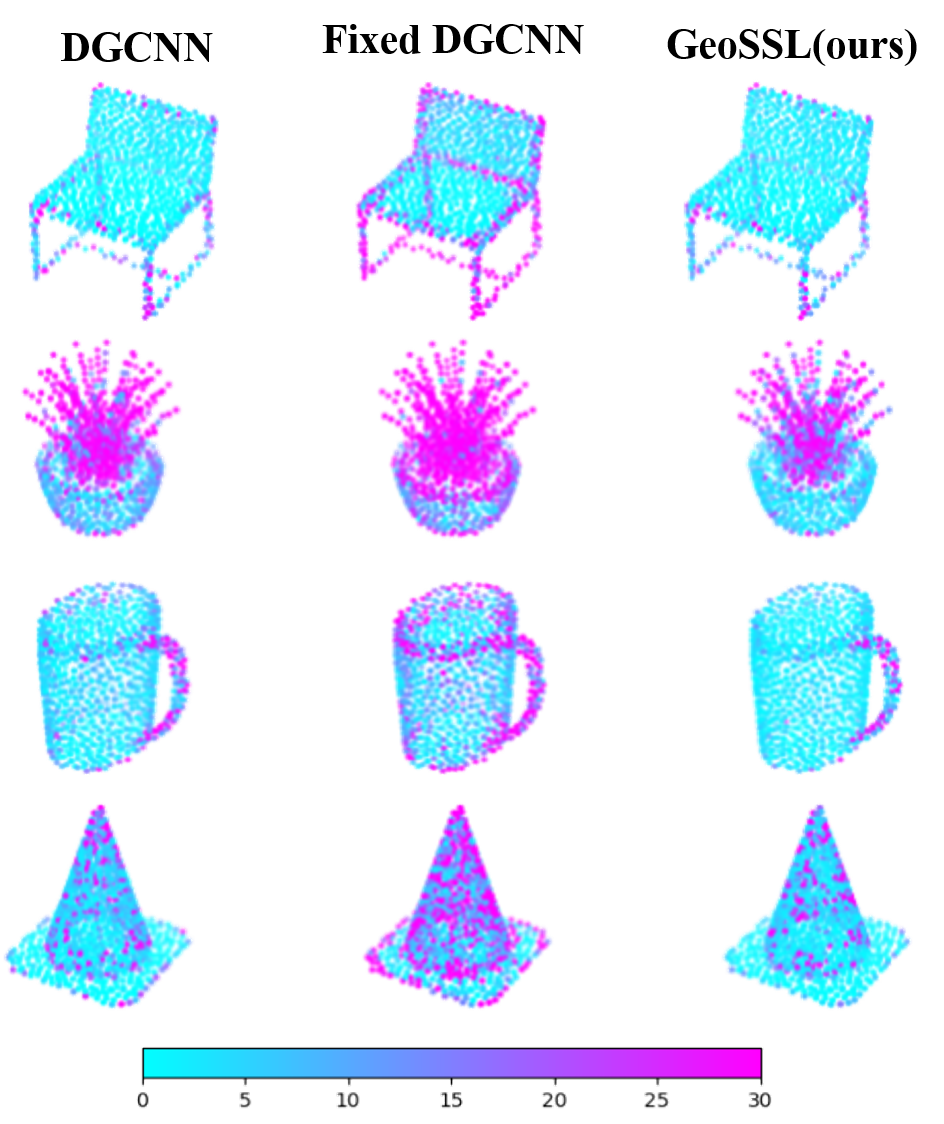}
\caption{Quantitative comparisons of normal estimation for GeoSSL$_\text{dgcnn}$ and DGCNN, the colors of points correspond to angular difference (estimation error) between predicted normal and ground truth normal, which are mapped to a heat-map ranging from 0-30 degrees. The small its value, the better.} \label{fig:normal_error}
\end{center}
\end{figure}

\begin{table}[t]\centering
\caption{Classification performance (\%) of GeoSSL with other baseline CNN models on the ModelNet40. }\label{table9}
\vspace{0.1cm}	
\begin{centering}
\setlength{\tabcolsep}{7mm}
\begin{tabular}{l|c}
\hline
Methods & Overall Accuracy  \\
\hline
Pointnet++ \cite{qi2017pointnet++}  & 90.7  \\
PointCNN \cite{li2018pointcnn}  & 92.2  \\
DGCNN+ \cite{wang2018dynamic}  & 92.2  \\
\hline
GeoSSL$_\text{pointnet++}$ 	 & 91.7  \\
GeoSSL$_\text{pointcnn}$ 	 & 92.8 \\
GeoSSL$_\text{dgcnn}$     & \textbf{92.9}  \\
\hline			
\end{tabular}
\end{centering}
\end{table}
		
\vspace{0.1cm}\noindent\textbf{Evaluation across CNN backbone models --} 
Evaluative results on different CNN baselines (\ie PointNet++ \cite{qi2017pointnet++}, DGCNN \cite{wang2018dynamic}, and PointCNN \cite{li2018pointcnn}) are illustrated in Table \ref{table9}. We can evidently find out that, our proposed methods can consistently outperform their baseline models.  
It further confirms that the nature of auxiliary geometric learning on improving semantic point cloud recognition.
		
\begin{table}[h]\centering
\caption{Comparisons of various of multi-task on the ShapeNet Part. }\label{table11}\vspace{0.1cm}
\begin{centering}
\setlength{\tabcolsep}{3mm}
\begin{tabular}{l|c|c}
\hline	
Methods & Classification &  Segmentation  \\
&  Accuracy(\%) & IoU(\%) \\
\hline
DGCNN+ \cite{wang2018dynamic} & 98.8 & 85.1 \\
\hline
GeoSSL$_{cls+seg}$  & 98.9 & 84.4 \\
\hline
GeoSSL$_{cls+reg}$  & \textbf{99.4} & - \\
\hline
GeoSSL$_{seg+reg}$  & -& \textbf{85.7} \\
\hline			
\end{tabular}
\end{centering}
\end{table}

\vspace{0.1cm}\noindent\textbf{Evaluation on multi-task learning architecture --} To this end, we additionally conducted experiments on the ShapeNet Part dataset. 
The network architecture used here is the same as in Fig. \ref{fig:GPNet}, the only difference lies in the task setting. 
Comparison results are shown in Table \ref{table11}, where we evaluate different options of combining two tasks in a multi-task learning framework.
As can be seen from Table \ref{table11}, when simply combining classification and segmentation tasks in a multi-task manner, denoted as MTNet$_{cls+seg}$. The classification performance (98.9\%) of the MTNet$_{cls+seg}$ is only slightly better than its baseline DGCNN (98.8\%), but even worse than its baseline DGCNN on segmentation performance (84.4\%). Different from that, our models with an auxiliary fitting on geometric properties achieve superior results to the DGCNN and MTNet$_{cls+seg}$ both on classification (GeoSSL$_{cls+reg}$) and segmentation (GeoSSL$_{cls+seg}$) tasks, which further demonstrates performance gain of our method can be credited to additional regression learning branch. 



\vspace{0.1cm}\noindent\textbf{Evaluation on estimation of geometric properties --} 
Fig. \ref{fig:normal} and \ref{fig:cur}  visualize the predicted normals and curvatures with the proposed GeoSSL respectively. 
From which we can see that estimation performance of our method are very close to the ground truth. 
Furthermore, when neural networks are trained on clean point sets, they could predict more accurate normals than those obtained by geometric computation, especially for noisy testing set. 
This could be attributed to their capability to learn statistical regularities from training data. 
For verification, we train a DGCNN based normal estimation network using clean training sets of $1024$ points from the ModelNet40; for testing, we add Gaussian permutations to $2468$ instances of point sets, where the noise level for each point is $\sigma = 0.01$ (clean point sets are normalized in a unit sphere). 
Geometric computation produces an averaged error of $1.0015$ against GT normals (measured in the Cosine distance, ranging in $[0, 2]$), and our trained neural model gives a lower one of $0.7332$, which verifies our claim the learning based method with clean data can predict more accurate geometric properties.

\begin{table}[h]
\caption{Effect of different $\lambda$ proportion of two loss in GeoSSL$_\text{dgcnn}$. The smaller $\lambda$ is, the less effect of local geometric learning affects.   }\label{table13}\vspace{0.1cm}
\setlength{\tabcolsep}{1.8mm}
\centering
\begin{tabular}{l|c|c|c|c|c|c}
\hline
Setting ($\lambda$ ) &1 & e-1 & e-2 & e-3 & e-4 & e-5 \\
\hline
Accuracy (\%) & 87.1 & 89.3 & \textbf{ 92.9} & 92.3 &91.9&91.7 \\
\hline
\end{tabular}
\end{table}		

\vspace{0.1cm}\noindent\textbf{Evaluation on ratio between losses --} In our classification settings, $\lambda$ is an important parameter to determine the proportion of two loss function (\ie the regression loss for fitting local geometries and the classification/segmentation loss).  
We hold out 20\% of training data as the validation set. 
We observe that the trade-off parameter $\lambda$ varies across different network architectures and different tasks, but when $\lambda$ is set as between [$10^{-3}$, $10^{-2}$], our model can steadily perform well. 
As a result, we select either 0.01 or 0.001 for $\lambda$ in our experiments. 
Specifically, Table \ref{table13} illustrates the trend of classification accuracy with $\lambda$ varying on the ModelNet40 with GeoSSL$_\text{dgcnn}$. 
When $\lambda$ = e-2, it can reach the best classification performance.
	
\begin{table}[h]
\caption{Evaluation on transferring knowledge for 3D classification using the GeoSSL$_\text{dgcnn}$.}\label{table12}\centering
\setlength{\tabcolsep}{6mm}
\begin{tabular}{l|c}
\hline
Experiment Setting  & Accuracy \\
\hline
Random initialization & 92.5 \\
Pre-trained on the ShapeCore   & \textbf{92.9} \\
\hline
\end{tabular}	
\end{table}

\vspace{0.1cm}\noindent\textbf{Effects of pre-training with auxiliary data --} An experiment to evaluate the effects of auxiliary data on pre-training is conducted by pre-training the proposed models on the ShapeNetCore dataset.
Results in Table \ref{table12} show that moderate improvement on the pre-trained models can be achieved over the identical network with random initialization, which encourages us to adopt pre-training for boosting performance.

\section{Conclusion}
This paper, for the first time, systematically introduces self-supervised learning into 3D point cloud semantic analysis, which is a generic method to readily replace its backbone with any other deep geometric learning. 
Rather than employing geometric properties as additional feature input, our network utilizes them as auxiliary supervision signals, which can consistently improve performance on semantic analysis. Given accurate privileged local shape information, our method can further be boosted to 93.5\% mean classification accuracy on the ModelNet40.

		
		
\ifCLASSOPTIONcaptionsoff
\newpage
\fi
\bibliographystyle{IEEEtran}
\bibliography{IEEEabrv,GeoSSL}
		

\end{document}